\pdfoutput=1
\documentclass[11pt]{article}

\usepackage{emnlp2021}

\usepackage{times}
\usepackage{latexsym}
\usepackage{amssymb}
\usepackage{graphicx}
\usepackage{enumitem}
\usepackage{paralist}
\usepackage{amsmath}
\usepackage{algorithm}
\usepackage{algorithmic}
\usepackage{bbding}
\usepackage{pifont}
\usepackage{wasysym}

\usepackage[T1]{fontenc}
\usepackage[utf8]{inputenc}

\usepackage{microtype}
\usepackage{newfloat}
\usepackage{listings}
\usepackage{color}
\usepackage{soul}
\usepackage{multirow}
\usepackage{makecell}
\usepackage[hang,flushmargin]{footmisc}

\newcommand{\STAB}[1]{\begin{tabular}{@{}c@{}}#1\end{tabular}}

\title{StoryER: Automatic Story Evaluation via Ranking, Rating and Reasoning}

\author{Hong Chen$^{1,3}$,  Duc Minh Vo$^1$,  Hiroya Takamura$^{2,3}$,  Yusuke Miyao$^{1,3}$, Hideki Nakayama$^{1,3}$\\ The University of Tokyo$^1$, Tokyo Institute of Technology$^2$\\ National Institute of Advanced Industrial Science and Technology, Japan$^3$  \\  \texttt{\{chen, vmduc, nakayama\}@nlab.ci.i.u-tokyo.ac.jp}\\\texttt{yusuke@is.s.u-tokyo.ac.jp}\\\texttt{takamura.hiroya@aist.go.jp}
}

\begin{document}
\maketitle
\setlength{\abovedisplayskip}{3pt}
\setlength{\belowdisplayskip}{3pt}
\begin{abstract}

Existing automatic story evaluation methods place a premium on story lexical level coherence, deviating from human preference.
We go beyond this limitation by considering a novel \textbf{Story} \textbf{E}valuation method that mimics human preference when judging a story, namely \textbf{StoryER}, which consists of three sub-tasks: \textbf{R}anking, \textbf{R}ating and \textbf{R}easoning.
Given either a machine-generated or a human-written story, StoryER requires the machine to output 1) a preference score that corresponds to human preference, 2) specific ratings and their corresponding confidences and 3) comments for various aspects (e.g., opening, character-shaping).
To support these tasks, we introduce a well-annotated dataset comprising (i) 100k ranked story pairs; and (ii) a set of 46k ratings and comments on various aspects of the story.
We finetune Longformer-Encoder-Decoder (LED) on the collected dataset, with the encoder responsible for preference score and aspect prediction and the decoder for comment generation.
Our comprehensive experiments result in a competitive benchmark for each task, showing the high correlation to human preference.
In addition, we have witnessed the joint learning of the preference scores, the aspect ratings, and the comments brings gain in each single task.
Our dataset and benchmarks are publicly available to advance the research of story evaluation tasks.\footnote{Dataset and pre-trained model demo are available at anonymous website \url{http://storytelling-lab.com/eval} and \url{https://github.com/sairin1202/StoryER}}

\end{abstract}
\section{Introduction}

\begin{figure}[tb]
    \centering
    \includegraphics[scale=0.48]{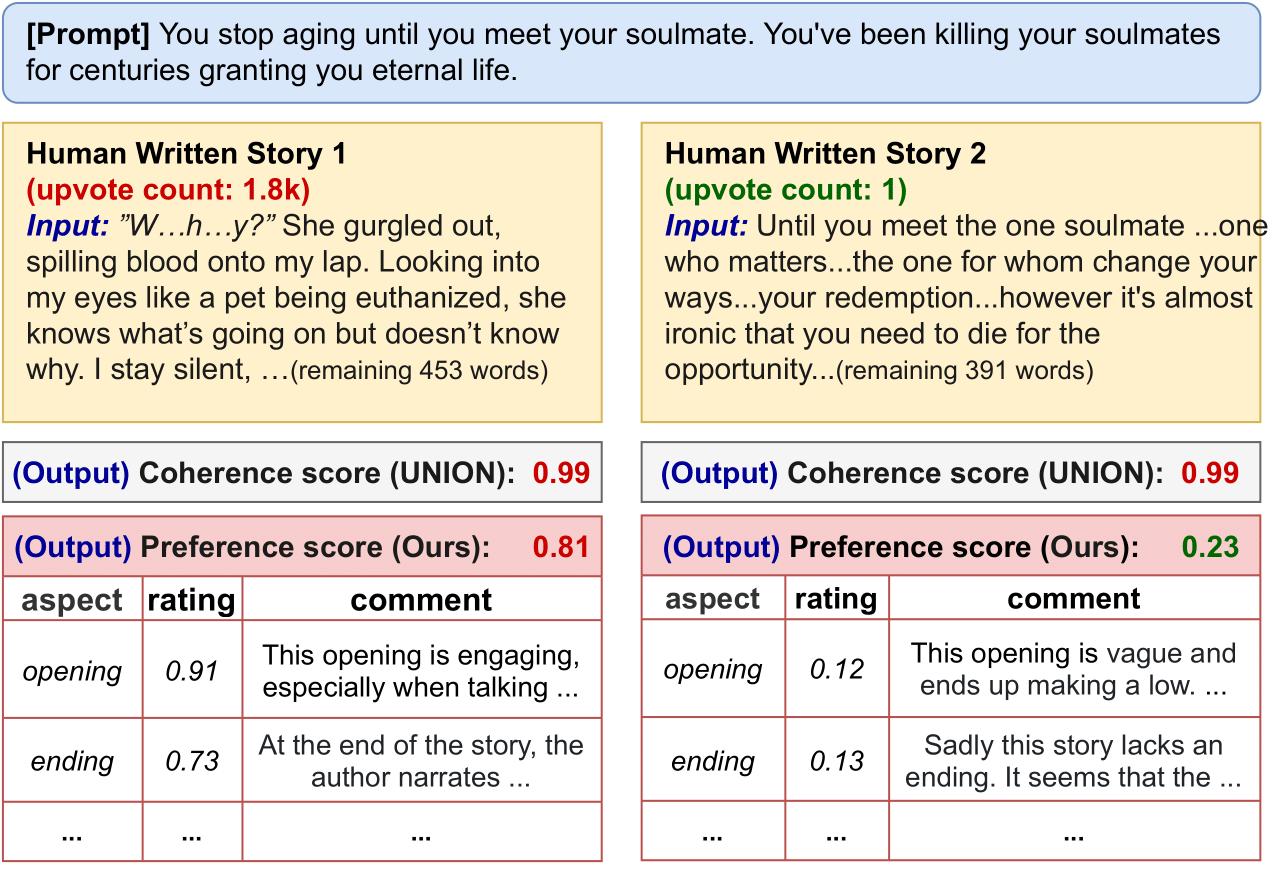}
    \caption{The existing story evaluation method (UNION) outputs a score for estimating the coherence of the stories, while human-written stories rarely suffer from this problem. Our model (Ours) which is trained by comparing two stories (Ranking), evaluates the story based on human preference (i.e., upvote counts), produces scores for various aspects (Rating), and leaves comments (Reasoning). Our model is applicable to both machine-generated and human-written stories.}
    \label{fig:overall}
    \vspace*{-1\baselineskip}
\end{figure}

Even for humans, evaluating story quality is a challenging task. Although many literature criteria have been proposed, the most straightforward way is to count how many readers like the story which is referred as to human preference.
Bearing it in mind, story writing community usually uses upvote count as a story quality criterion.
% As the consensus on story writing community, the upvote count serves as a criterion of story quality.
As shown in Fig.~\ref{fig:overall}, more readers like the left story (upvote count = 1.8k) rather than the right one (upvote count = 1).

% Despite the success in evaluating machine-generated stories, \textit{referenced metrics} (e.g., BLEU~\cite{papineni2002bleu}, METEOR~\cite{banerjee2005meteor}, ROUGE~\cite{lin2004rouge}) and \textit{unreferenced metrics} (e.g., UNION~\cite{guan2020union}, MANPLTS~\cite{ghazarianplot}) are struggled with human-written stories, as illustrated in Fig.~\ref{fig:overall}.
% This is primarily due to the fact that such methods emphasize word overlapping, text coherence, and consistency issues (e.g., repeated plots, long-term consistency), which do not commonly occur in human-written stories.
% As a step toward developing story evaluation in agreement with human preference regardless of whether the story is generated by a machine or written by a human, we present a novel task of \textit{preference-aware story evaluation}.
% Despite the success in evaluating story lexical level coherence, the scores from 

Existing methods which use 
\textit{referenced metrics} (e.g., BLEU~\cite{papineni2002bleu}, METEOR~\cite{banerjee2005meteor}, ROUGE~\cite{lin2004rouge}) and \textit{unreferenced metrics} (e.g., UNION~\cite{guan2020union}, MANPLTS~\cite{ghazarianplot}), deviate from human preference (Fig.~\ref{fig:overall}).
% , as illustrated in Fig.~\ref{fig:overall}. 
On the contrary, we aim to explicitly evaluate a story, introducing a human preference-liked system consisting of three subtasks: Ranking, Rating and Reasoning.
% To enable explicit story evaluation, in this paper we propose \textbf{StoryER}, a story evaluation system consisting of three subtasks: Ranking, Rating and Reasoning.
% Technically, inspired by human evaluation, we train a model with the ranking objective function, outputting a \hl{preference} score in the inference time (Ranking).
% Concurrently, to further explain the evaluation, the model gives specific scores (Rating) and generates comments (Reasoning) for various story aspects such as opening, character shaping.
% We believe that this novel task will close the performance gap between automatic evaluation and human preference, leading to a more general evaluation of stories.

% Inspired by pairwise human evaluation, we often request the annotators to compare two stories generated by the same inputs (e.g., prompt). Our model is trained using the ranking objective function and may output a score in the inference time (Ranking).
% For further evaluation explanation, we require our model to evaluate the scores (Rating) and generate comments (Reasoning) for each predefined aspect.
% % Therefore, in this paper we propose \textbf{StoryER}: Story evaluation via Ranking, Rating and Reasoning ,which thoroughly evaluate the story invoked by a comprehensive understanding of the story.
% We believe that this novel task will close the performance gap between automatic evaluation and human preference, leading to a more general evaluation of stories.

\begin{figure*}
    \centering
    \includegraphics[scale=0.4]{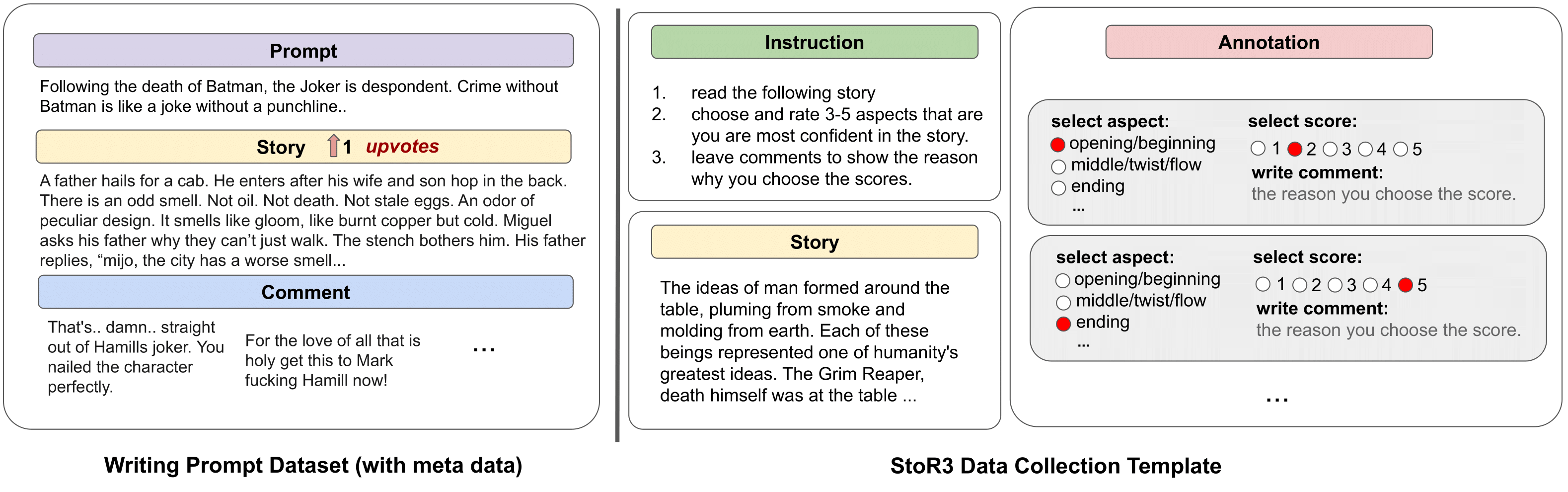}
    \caption{The Writing Prompt Dataset with metadata (left) contains prompt, story, upvotes, and comments from readers. Our dataset collection pipeline (right) shows the template for data collection. We ask the workers to select 3-5 aspects, score each aspect 1-5 from poor to good and leave the comments that shows the reason for the score they rated.}
    \vspace*{-1\baselineskip}
    \label{fig:dataset}
\end{figure*}

% \hl{what is this data used for? 100k is used to train and evaluate the preference score prediction.}
% We ask workers from Amazon Mechanical Turk (AMT) to annotate 28,099 comments and scores for 9,122 (of 63,929) stories (see data collection template in Figure~\ref{fig:dataset} (right)).
% Whilst the rest is augmented data derived from filtering the noisy and uncategorized comments in WP using an aspect classification and sentiment analysis model trained on our collected 28,099 comments and scores.

We build a model upon Longformer-Encoder-Decoder (LED)~\cite{beltagylongformer}, where the encoder predicts the preference score (Ranking), aspect ratings and confidences (Rating) while the decoder generates the comments (Reasoning).
Inspired by widely-used pairwise comparison in story evaluation, we train our model with the ranking objectives.
In this way, the score margin between good and poor stories are enlarged, resulting in high correlation between human preference and our predicted preference score (Fig.~\ref{fig:overall}).
We also witness that our performance is improved when we conduct joint training on three subtasks.

In aid of the proposed task, we present a well-annotated crowd-sourcing dataset, consisting of two parts.
(i) One is built from 63,929 stories and their corresponding upvote counts provided in WritingPrompt dataset (WP)~\cite{fan-etal-2018-hierarchical} (Figure~\ref{fig:dataset} (left)) by pairing one highly-upvoted story (upvotes $\geq$ 50) and one lowly-upvoted story (upvotes $\leq$ 0) under the same prompt.
As a result, we obtain 100k pairs of stories, namely \textit{100k story ranking data}, used to train and evaluate the preference score prediction.
(ii) The other part is made up of 45,948 aspect comments and their respective rating scores (1-5) by Amazon Mechanical Turk (AMT) and augmented data (Section~\ref{sec:46k}), namely \textit{46k aspect rating and reasoning data}, used for model explanation.
Our contributions are three-fold:

% , which is widely used in human evaluation in story generation tasks, we employ ranking objectives to train our model.
% In this way, the model enlarges the score margin between good and poor stories, leading to a high correlation to human preference in our preference score prediction, as shown in Fig.~\ref{fig:overall}.
% We also witness that our performance improves when we conduct joint training on three subtasks.
% Moreover, our model can be extended into a hybrid version that takes into account story coherence issues by incorporating negative stories from previous works. 
% Our contributions are three-fold:

\begin{compactitem}
    \item This study addresses a novel task StoryER, that consists of preference score prediction, aspect rating and comment generation.
    % , this paper is the first developing automatic system on evaluating the story by its content.
    \item We introduce a new dataset for StoryER task and create benchmarks to promote the story evaluation research. 
    % \footnote{For review process, dataset and pre-trained model are available at anonymous website \url{http://storytelling-lab.com/eval}}
    % Dataset and pre-trained model are available in  \url{http://storytelling-lab.com/eval}\footnote{This website is anonymous for double-blind reviewing at the time of submission.}.
    \item Comprehensive experiments and intensive analysis indicate our preference score prediction outperforms previous metrics, and more accurately reflects human preference. Aspect rating and comment generation also helps in the evaluation and provide explanations. Moreover, we point out the remaining challenges under various scenarios in the hope that facilitates future research.  
    
    % Compared with previous works, our model yields significantly higher performance in the evaluation of content.  We perform ablation tests and comprehensive analysis to demonstrate the effectiveness of the proposed model and dataset. We also conduct experiment to show that this task remains challenging in some scenario, thus requiring more focused future research on stories in different domains.
\end{compactitem}

% \begin{figure}
%     \centering
%     \includegraphics[width=0.3\textwidth]{LaTeX/figure/overall.pdf}
%     \caption{Caption}
%     \label{fig:my_label}
% \end{figure}
\section{Related work}
% Existing automatic story evaluation methods can be roughly divided into overlap-based metrics, neural-based metrics and neural discriminators based on their scoring methods, which we briefly summarize as follow. 

\noindent \textbf{Overlap-based metrics} such as BLEU~\cite{sulem2018bleu} and ROUGE~\cite{lin2004rouge} calculate lexical matches (i.e., n-gram matching) and reward the words that resemble the reference in their surface form, even if they do not accurately capture meaning, and penalize other paraphrases. 
Recent research~\cite{edunov2020evaluation} indicates that these metrics do not reflect human preferences, particularly for open-ended text generation tasks.

\noindent \textbf{Neural-based metrics} are motivated by the success of transformers as multitask learners~\cite{vaswani2017attention}, and adapt them for the task of neural language evaluation.
When compared to overlap-based metrics, BERTScore~\cite{zhang2019bertscore}, MoverScore~\cite{zhao2019moverscore}, BLEURT~\cite{sellam2020bleurt} report stronger correlations with human judgment.
For specific use, in open dialogue generation, Adem~\cite{lowe-etal-2017-towards} captures semantic similarity beyond word overlap statistics, and exploits both the context and the reference response to calculate its score for the model response.
RUBER~\cite{tao2018ruber} and its variant, RUBER-BERT~\cite{ghazarian2019better} evaluates a reply by taking into consideration both a ground-truth reply and a query without requiring labels of human satisfaction and can be extended to different datasets and languages.

\noindent \textbf{Neural discriminator} is proposed particularly for story evaluation. The metrics mentioned above show limited performance in story evaluation as demonstrated in ~\citet{guan2021openmeva}.
UNION~\cite{guan2020union} and MANPLTS~\cite{ghazarianplot} analyze the problem from machine-generated stories and generate negative data by heuristics and plot manipulation, and then distinguish by a BERT-based model~\cite{devlin-etal-2019-bert}. The coherence score they produce can be expressed as the probability of the story being identified as human-written story.
In this paper, we require our model to follow human preference, not only the coherence, which we believe is a more general way of story evaluation.

\begin{table}[]
% \footnotesize
\centering
\resizebox{\linewidth}{!}{
\begin{tabular}{l|cccc}
\hline
      & \#prompt & \#$S_{high}$ (word) & \#$S_{low}$ (word) & \#pairs\\ \hline
Train & 5892        & 10371 (491.01)                   & 26246 (453.06)                 & 66336           \\
Val   & 2280        & 3816 (473.27)                    & 11458 (446.40)                  & 27748                \\
Test  & 2280        & 3906 (488.32)                    & 8132 (454.87)                   & 22887     \\\hline              
\end{tabular}
}
\caption{Data statistics of 100k story ranking data. \#prompt denotes the number of unique prompts, \#$S_{high}$ and \#$S_{low}$ denotes the number of highly-voted stories and lowly-voted stories. We also show the averaged word count in the parentheses. \#pairs shows the number of ranked story pairs.}
\vspace*{-1\baselineskip}
\label{tab:overall_score_dataset}
\end{table}

\begin{table*}[]
\footnotesize
\centering
\resizebox{\linewidth}{!}{
\begin{tabular}{l|cccccc}
\hline
                      & \#comment & \#comment* & rate (1-5) & rate* (1-5) & comment\_len & comment\_len* \\
\hline
\textbf{Structure} &  &   &    & &  &        \\
\quad opening/beginning     & 3615         & 5617          & 2.53       & 3.15        & 30.20        & 32.44         \\
\quad middle/twist/flow/conflict     & 3967         & 5971          & 2.24       & 2.78        & 30.59        & 31.63         \\
\quad ending                & 5610         & 7615          & 2.13       & 2.49        & 31.48        & 31.59   \\
\hline
\textbf{Writing Style} &  &   &    & &  &        \\
\quad character shaping     & 5101         & 7102          & 2.21       & 2.53        & 31.57        & 34.23         \\
\quad scene description     & 4168         & 6172          & 2.18       & 2.53        & 31.75        & 39.30         \\
\hline
\textbf{Type} &  &   &    & &  &        \\
\quad heartwarming/touching (Romance)    & 426          & 1866          & 2.99       & 4.39        & 32.05        & 32.64         \\
\quad sad/crying/tragedy (Tragedy)       & 462          & 1680          & 3.12       & 3.93        & 30.85        & 34.67         \\
\quad horror/scary (Horror)          & 815          & 1985          & 2.49       & 3.61        & 30.92        & 33.24         \\
\quad funny/hilarious/laugh (Comedy) & 1153         & 3156          & 3.25       & 3.96        & 30.04        & 30.91         \\
\quad novelty/good idea/brilliant (Fiction)            & 2782         & 4784          & 2.49       & 3.26        & 32.51        & 32.70         \\\hline
\bf{Overall}               & 28099        & 45948         & 2.56       & 3.26        & 31.20        & 33.33   \\\hline     
\end{tabular}}
\caption{Data statistics in 46k aspect rating and reasoning data. * denotes the data statistics after data augmentation. We list the number of comments with rating scores (2nd and 3rd columns), averaged rating scores (4th and 5th columns) and averaged word count (6th and 7th columns).}
\vspace*{-1\baselineskip}
\label{tab:comment_statistics}
\end{table*}

\section{Dataset}
Our dataset comprises of two parts: 100k story ranking, and 46k aspect rating and reasoning.\footnote{All data collection follows the same procedure as described in the previous work~\cite{fan-etal-2018-hierarchical} on Reddit, which comply with ACL Code of Ethics.}

\subsection{100k Story Ranking Data}
% \hl{why using pairwise data (added)}
% \hl{how many pairs do we obtain finally? revised}
As we mentioned above, ranking method is more flexible and better than discrimination when evaluating the story (we also experimentally compare them in Sec.~\ref{sec:rank}).
We thus prepare 100k pairwise ranking data for training the model.
To this end, we first collect 193,842 stories prior to 03/2020 from WP\footnote{\url{https://huggingface.co/datasets/rewardsignal/reddit_writing_prompts}} along with their prompt, the number of upvotes and uncategorized comments.
%To prevent the risk of a good story receiving few upvotes since it is newly updated, we remove the stories updated from 12/2019 to 03/2020.
We remove the stories updated from 12/2019 to 03/2020, since newly-updated stories usually have few upvotes regardless of whether they are good or bad.
% To avoid the possibility of an excellent story with few upvotes because it is newly updated, we remove the stories from 12/2019 to 03/2020.
Then, we exclusively keep stories with the word count between 200 and 800.
Finally, we pick two stories from the same prompt, one highly upvoted (i.e., upvotes $\geq$ 50~\footnote{we notice that some stories that receive upvotes $\geq$ 50 can be listed in /r/bestofWritingPrompts/}) and one lowly upvoted (i.e., upvotes $\leq$ 0), resulting in a total of 63,929 unique stories and 116,971 story pairs.
We split the story pairs based on the prompts into training, validation and testing (Table~\ref{tab:overall_score_dataset}), to ensure that each division receives a unique set of prompts. 
% Table ~\ref{tab:overall_score_dataset} summarizes the statistics in the dataset that were used to train the overall score prediction via ranking.

% In this task, we apply ranking instead of discrimination as its high flexibility and performance. 

% \chen{not fluency} We put a discussion of comparing two methods in Sec.~\ref{sec:rank}.
% Our data extracted from WP with meta data \footnote{\url{https://huggingface.co/datasets/rewardsignal/reddit_writing_prompts}} that
% contains 193842 stories prior to 03/2020, each of which has its prompt, the number of upvotes and uncategorized comments.
% We begin with removing the stories from 12/2019 to 03/2020, in order to avoid the possibility that an excellent story would receive few upvotes simply because it is new.
% Then, we exclusively keep stories with the word count between 200 and 800, in order to exclude very short and long stories.
% We pick up two stories with one highly-upvoted story (i.e. upvotes $\geq$ 50) and one lowly-upvoted story (i.e. upvotes $\leq$ 0) from the same prompt, results in a total of 63929 unique stories and 116971 story pairs.
% We split the story pairs based on the prompts into training, validation and testing, to ensure that each division received a unique set of prompts. Table ~\ref{tab:overall_score_dataset} summarizes the statistics in the dataset that were used to train the overall score prediction via ranking.

\subsection{46k Aspect Rating and Reasoning Data} \label{sec:46k}
Apart from the preference score, we require our model to provide ratings and comments on pre-defined aspects to aid in the explanation of the predicted preference score.

\noindent
\textbf{Aspect category extraction.}
To begin with, we must determine which aspects in the content should be measured.
As some readers leave comments to explain why they upvote or downvote the stories, a straightforward way is to extract aspect categories based on those uncategorized comments.
We therefore adopt latent Dirichlet allocation (LDA), which models the documents with a certain number of topics, based upon the co-occurrence of individual words.
More precisely, we follow~\citet{brody-elhadad-2010-unsupervised} to treat each comment as a separate document. 
LDA can produce a distribution of frequency of occurrence for each word in the topics. 
We optimize LDA through a cluster validation scheme, and obtain the optimal number of aspects 10.
Based on the most representative words in each topic, we manually name each topic as the aspect category.
These aspect categories are defined using some widely used aspects inspired from the websites.\footnote{list in the supplementary material}

% Among the unsupervised approaches, the common choices are latent Dirichlet allocation (LDA) based methods~\cite{titov-mcdonald-2008-joint, brody-elhadad-2010-unsupervised, mukherjee-liu-2012-aspect}. 
% Following ~\citet{brody-elhadad-2010-unsupervised}, we use a standard implementation of LDA, and treat each sentence as a separate document. The output of the model is a distribution over inferred aspects for each sentence in the data. We run the algorithm and employed a cluster validation scheme that they proposed. Then we obtain the the optimal number of aspects as 10 \hl{why?(revised)} and manually annotate each cluster based on the most representative words. The aspect categories are defined using the well-used aspects that collected from the websites~\footnote{list in the supplementary material}.
% The final aspects are listed in Table~\ref{tab:comment_statistics}.
% \hl{is there any citation to support the aspect category? (revised)}

\noindent
\textbf{Comment and aspect collection.}
Comments in WP meta data are neither categorized with aspect categories, nor labeled with ratings, and some of them are totally irrelevant to the content. More importantly, there is a bias towards positive comments, which implies that not too many readers are willing to leave comments on poor stories. Therefore, we collect new comments via crowd-sourcing. By learning from these well-annotated comment data, we train neural models to filter out noisy data from comments in WP meta data.
To collect the data, we ask workers from AMT to select aspects, rate sentiment and leave comments on 5,964 unique stories from WP. 
For increasing the diversity of comments, some stories are allocated to two different annotators, resulting in a total of 9,112 submissions (i.e., 1.53 annotations/story). As shown in Figure~\ref{fig:dataset} (right), each story requires the annotators to rate (normalized to 0-1) and leave comments on 3 to 5 aspects that are most confident by the workers. The final statistics of the comments is listed in Table~\ref{tab:comment_statistics}. 
% To note that, during the annotation, we also show the upvote counts to the workers, and ask them to find positive reasons for highly-upvoted stories and negative reasons for lowly-upvoted stories.

% \hl{as discussed above, most of comments are positive. What is distribution of aspect rating after augmentation? I suppose all comments get high scores. Moreover, do you analyze the agreement between story (overall score) and comments? Say, the story with high score should have good comments, and vice versa.}\chen{it can be seen in Table 2 rate}

\noindent \textbf{Comment augmentation.}
The noisy comments in WP meta data then can be classified and analyzed by two models: aspect category classification model and comment sentiment analysis model that trained with our collected data. 
The training details can be found in the supplementary material.
We filter out irrelevant comments by eliminating those with no values in aspect categories that exceeds 0.9 after softmax and retain the comments with the word count ranged from 15 to 50.
The remaining comments are then rated by the their sentiments.
Finally, we obtain 17,849 valuable comments for 6,705 additional unique stories and merge them into our collected data, resulting in a total number of 45,948 for comments and 12,669 for unique stories. 
We split the collected data into training, validation, and test data in the ratio of 8:1:1 and put the augmented data into the training data (Table~\ref{tab:comment_statistics}).

% With the aspect category and sentiment rating, these comments and corresponded stories can be merged into our collected data.
% To this end, our augmented data, including collected data, contains 45948 comments and 12669 unique stories. We split the collected data into training, validation, and test data in the ratio of 8:1:1 and put the augmented data into the training data.
% We list the statistics of the augmented comments in Table~\ref{tab:comment_statistics}. Noticing that, after augmentation, the averaged sentiment ratings increases a lot that proves that the comments in the WP meta data are biased to positive.

\section{StoryER}
\subsection{Task Definition}

% \chen{need double check}
Given a story $\mathbf{s}$, the task is to output a set $\left \{ p_s, \mathbf{a}^{\rm c}, \mathbf{a}^{\rm r}, \mathbf{c} \right \}$ where $p_s$ denotes the preference score of the story $\mathbf{s}$, which is used for comparing story quality.
For more explicit explanation, we further output confidence scores
$\mathbf{a}^{\rm c} = \left \{ a^{\rm c}_{k}\right \}^{K}_{k=1}$, aspect ratings
$\mathbf{a}^{\rm r} = \left \{ a^{\rm r}_{k}\right \}^{K}_{k=1}$, and comments
$\mathbf{c} = \left \{ c_{k}\right \}^{K}_{k=1}$
 for $K$ aspects ($K=10$ in our experiments), respectively.
Confidence scores $\mathbf{a}^{\rm c}$ reflect the likelihood of utilizing the specific aspects as measures, as some aspects (e.g., horror) are not applicable in some stories (e.g., comic story).
Aspect ratings $a^{\rm r}_{k}$ are considered as the scores of each aspect.
Comments $\mathbf{c}$ demonstrate the reason that the reader upvotes/downvotes the story, producing a more explicit explanation for the aspect rating.
%In our following experiment, 
%we assume that $\sum_{k=1}^{K}a^{\rm c}_{k}=1$ is true for aspect confidence, and $a^{\rm r}_{k} \in [0, 1]$ is true for aspect rating calculated separately during the training.
We assume $\sum_{k=1}^{K}a^{\rm c}_{k}=1$ for aspect confidence, and $a^{\rm r}_{k} \in [0, 1]$ for aspect rating, which is calculated separately during the training.
%we define the aspect confidence $\sum_{k=1}^{K}a^{\rm c}_{k}=1$, and each aspect rating $a^{\rm r}_{k} \in [0, 1]$ calculated separately during the training. 
% \hl{change this statement}

Please note that aspect rating and comment generation results are not used as metrics in this work, while they are used for 1) improving preference score prediction by joint learning, and 2) producing explanation. Investigating how to include them into metrics is a future direction for this research.

\subsection{Learning a Story Evaluator}
% \hl{Do we need this sentence?}
% Recent work in constructing large-scale generative language models based on transformers~\cite{vaswani2017attention} has led to considerable improvements in natural language understanding and generation tasks.

\begin{figure}
    \centering
    \includegraphics[scale=0.44]{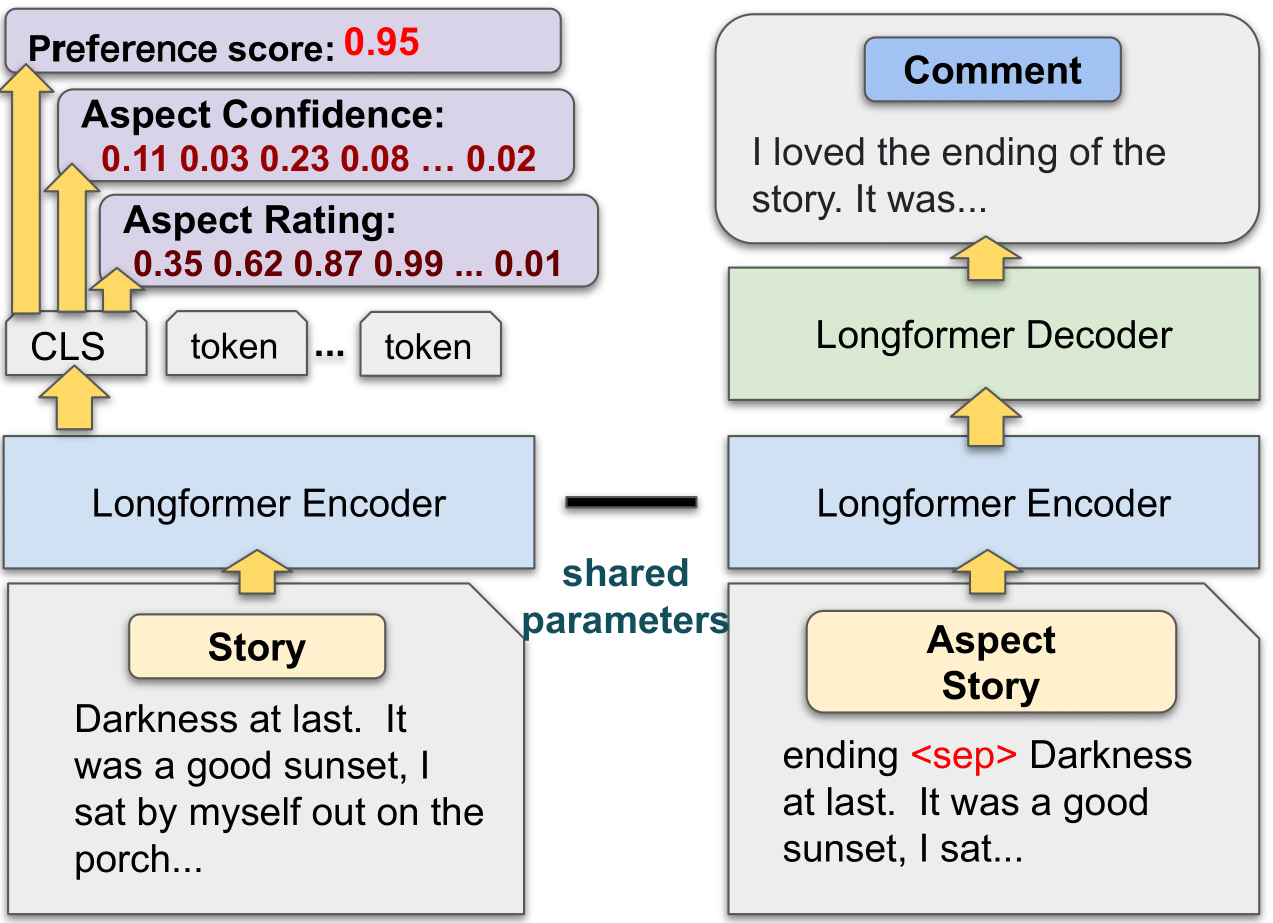}
    \caption{Overview of our model. The encoder (left) predicts the preference score, aspect confidence, and aspect rating. The decoder (right) generates the comment for each aspect.}
    \vspace*{-1\baselineskip}
    \label{fig:model}
\end{figure}

Following~\citet{ghazarianplot}, we use Longformer-Encoder-Decoder (LED)~\cite{beltagylongformer} to produce a preference score, as well as ratings and comments for the pre-defined aspects. As shown in Figure~\ref{fig:model}, we encode the story $\mathbf{s}$, and use its feature on the special token (i.e., [CLS]) to predict the preference score $p_s$, aspect confidence $\mathbf{a}^{\rm c}$ and rating $\mathbf{a}^{\rm r}$ by additional layers. For generating comments, we concatenate the story with aspect category name with a special token (i.e., $\textless$sep$\textgreater$), and send it into the same encoder. The decoder outputs the comment $\mathbf{c}$ that implies the performance of the story on the given aspect.

\subsection{Task 1: Preference Score Prediction (Ranking)}
Our model learns to predict the preference score by ranking two stories from the same prompt. 
As shown in Figure~\ref{fig:model}, we use the feature of [CLS] in the story, following a linear layer with sigmoid activation and finally turning it into a scalar score. 
We take Margin Ranking Loss to enlarge the margin gap $m$ of the scores between stories with high and low upvotes:
% \footnotesize
\begingroup\makeatletter\def\f@size{10}\check@mathfonts
\begin{equation}
    \mathcal{L}_{\rm p_s} =  \max(0, \sigma (\mathbf{W}_{\rm p_s} \mathbf{v}_{s_{low}}) - \sigma(\mathbf{W}_{\rm p_s} \mathbf{v}_{s_{high}} ) + m),
\end{equation}
\endgroup
% \normalsize
where $\mathbf{W}_{\rm p_s}$ denotes a linear layer for the feature of the story $\mathbf{v}_{s}$. $\sigma(\cdot)$ is the sigmoid activation function. $s_{high}$ and $s_{low}$ represent the highly-upvoted and lowly-upvoted stories.

\paragraph{Negative sample.}
Machine-generated stories often suffer from the coherence and consistency problem, while human-written stories usually do not. 
Therefore our model trained on human-written stories can hardly evaluate story coherence.
To enable our model to evaluate story considering coherence issues, we further train our model (Ours (N)) with negative stories that are generated by the methods in the previous works~\cite{guan2020union, ghazarianplot}.
We change the margin ranking loss as follow:
% \footnotesize
\begingroup\makeatletter\def\f@size{10}\check@mathfonts
\begin{align}
    \mathcal{L}_{\rm {pref}} &=  \max(0, \sigma (\mathbf{W}_{\rm p_s} \mathbf{v}_{s_{low}}) - \sigma(\mathbf{W}_{\rm p_s} \mathbf{v}_{s_{high}} ) + m),\nonumber\\
    \mathcal{L}_{\rm {coh}} &=  \max(0, \sigma (\mathbf{W}_{\rm p_s} \mathbf{v}_{s_{neg}}) - \sigma(\mathbf{W}_{\rm p_s} \mathbf{v}_{s_{low}} ) + m),\nonumber\\
    \mathcal{L}_{\rm p_s} &= \mathcal{L}_{\rm {pref}} + \mathcal{L}_{\rm {coh}},
\label{equ:neg}
\end{align}
\endgroup
% \normalsize
where $s_{neg}$ denotes the negative stories derived from the previous works.
In each iteration, we takes two pairs as training data: $s_{high}$ and $s_{low}$, $s_{low}$ and $s_{neg}$.

% \subsection{Joint Learning with Aspect, Comment and Negative Sample}
\subsection{Task 2: Aspect Confidence and Rating Prediction (Rating). }
We adopt two additional linear layers on the same feature $\mathbf{v}_s$ used in the story ranking. 
One is with learnable parameters $\mathbf{W}_{{\rm a}^{\rm c}}$, outputting confidence scores $\mathbf{a}^{\rm c} = \text{softmax}(\mathbf{W}_{\rm a^{\rm c}} \mathbf{v}_s)$.
The other one has $\mathbf{W}_{{\rm a}^{\rm r}}$, producing aspect rating $\mathbf{a}^{\rm r} = \sigma (\mathbf{W}_{\rm a^{\rm r}}\mathbf{v}_s)$.
Let $\mathbf{y_{{\rm a}^{\rm c}}} \in \left \{ 0, 1 \right \}^K$, $\mathbf{y_{{\rm a}^{\rm r}}} \in [0, 1]^K$ be the ground-truth confidence and rating, we define the confidence and rating loss functions as follows:
% One is used for aspect confidence prediction and the other is used for aspect rating.% Each outputs ten scalars which correspond to ten pre-defined aspects. The loss function is defined as follow: \hl{define notation} 
% \footnotesize
\begingroup\makeatletter\def\f@size{10}\check@mathfonts
\begin{align}
    % \mathbf{y_{{\rm a}^{\rm c}}} &\in \left \{ 0, 1 \right \}^K, \mathbf{y_{{\rm a}^{\rm r}}} \in [0, 1]^K, \mathbf{x_{{\rm a}^{\rm c}}} \in [0, 1]^K, \mathbf{x_{{\rm a}^{\rm r}}} \in [0, 1]^K, \nonumber\\
    % \mathbf{x_{{\rm a}^{\rm c}}} &= \text{softmax}(\mathbf{W}_{a^{\rm c}} \mathbf{v}_s),
    % \mathbf{x_{{\rm a}^{\rm r}}} = \sigma (\mathbf{W}_{a^{\rm r}}\mathbf{v}_s),\nonumber\\
    \mathcal{L}_{{\rm a}^{\rm c}} &= -\sum_{k=1}^{K} \mathbf{y}_{a^{\rm c}}[k]\log \mathbf{a}^{\rm c}[k],\\
    \mathcal{L}_{{\rm a}^{\rm r}} &= -\sum_{k \in M_s} \mathbf{y}_{\rm a^{\rm r}}[k] \log \mathbf{a}^{\rm r}[k]\\&   + (1 -  \mathbf{y}_{\rm a^{\rm r}}[k]) * \log (1 - \mathbf{a}^{\rm r}[k]).\nonumber
\end{align}
\endgroup
% \normalsize
% where $W_{a^{\rm c}}$ and $W_{a^{\rm r}}$ are two linear projection layers.
\noindent
We calculate the multi-class cross-entropy loss for the aspect confidence.
$\mathbf{y}_{\rm a^{\rm c}}[k] = 1$ if the $k$-th aspect is selected, otherwise $\mathbf{y}_{\rm a^{\rm c}}[k] = 0$.
For aspect rating, binary cross-entropy loss is calculated separately for each selected aspects.
$M_s$ denotes the set of aspects that are selected for story $\mathbf{s}$. 
$\mathbf{y}_{\rm a^{\rm r}}[k]$ denotes the normalized rating score for the $k$-th aspect. 

\subsection{Task 3: Comment Generation (Reasoning).}
The comments are generated conditioned on the aspect $\mathbf{a}$ and the story $\mathbf{s}$.
We input the concatenation of the aspect category name, special token, story, and train the LED under Maximum Likelihood Estimation (MLE) with the comment as target:
% \footnotesize
\begingroup\makeatletter\def\f@size{10}\check@mathfonts
\begin{equation}
    \mathcal{L}_{{\rm c}}\left(p_{\theta}\right)=- \sum_{t=1}^{\left|\mathbf{c}\right|} \log p_{\theta}\left({\mathbf{c}}_{t} \mid \mathbf{a}, \mathbf{s} , {\mathbf{c}}_{<t}\right),
\end{equation}
\endgroup
% \normalsize
where the $\mathbf{c}_t$ denotes the $t$-th token in the comment.

\noindent
For joint training three tasks, our final loss is the summation of all above loss functions:
% \footnotesize
\begingroup\makeatletter\def\f@size{10}\check@mathfonts
\begin{equation}
   \mathcal{L} =  \mathcal{L}_{\rm p_s} + \mathcal{L}_{{\rm a}^{\rm c}} + \mathcal{L}_{{\rm a}^{\rm r}} + \mathcal{L}_{\rm c}.
\end{equation}
\endgroup
% \normalsize

\subsection{Hyperparameters}
We conduct a comprehensive set of experiments to examine the effectiveness under different scenarios.
We fine-tune pre-trained LED from  Huggingface\footnote{\url{https://huggingface.co/transformers/model_doc/led.html}} with the batch size 16, the margin 0.3 and run 20k iterations for training (10 hours).
We adopt AdamW optimizer~\cite{loshchilov2018decoupled} with an initial learning rate of 4e-6, warming up in the first epoch and decreasing by a linear schedule. The reported results are averaged by the best results from three models with the same structure but initialized with three different seeds. 
More details and code can be found in the Appendix.

\begin{table*}[t]
\centering
\footnotesize
\resizebox{\linewidth}{!}{
\begin{tabular}{l|cc|cc|cc|cc|cc}
\hline
 &
  \multicolumn{6}{c|}{\textbf{Human Written Story}} &
  \multicolumn{4}{c}{\textbf{Machine Generated Story}} \\\cline{2-11}
 &
  \multicolumn{2}{c|}{Ranking} &
  \multicolumn{2}{c|}{WP$_{200}$} &
  \multicolumn{2}{c|}{SCARY$_{200}$} &
  \multicolumn{2}{c|}{PREF$_{200}$} &
  \multicolumn{2}{c}{COH$_{200}$} \\
Methods    & Acc(\%) & Dis    & $\rho$ & $\tau$ & $\rho$ & $\tau$ & $\rho$          & $\tau$          & $\rho$ & $\tau$ \\ \hline
PPL        & 61.80   & -      & 0.181* & 0.130* & 0.409* & 0.300*& -0.090 & -0.067  & -0.698*         & -0.452*         \\
ft-PPL     & 61.85   & -      & 0.168  & 0.120  & 0.449* & 0.328*& -0.098 & -0.074  & -0.690*         & -0.443*         \\
Ruber-bert & 39.16   & -0.032 & -0.019 & -0.014 & 0.032  & 0.026& -0.140 & -0.100  & -0.141          & -0.092           \\
UNION      & 48.74   & 0.003  & 0.001  & 0.001  & 0.156* & 0.114 & -0.068 & -0.045  & 0.188*          & 0.132*           \\
MANPLTS    & 53.08   & 0.016  & 0.124  & 0.107  & -0.070 & -0.061 &   0.164  & 0.130      &   0.729*          & 0.498*  \\
Ours &
  \textbf{73.93} &
  \textbf{0.228} &
  \textbf{0.583*} &
  \textbf{0.422*} &
  \textbf{0.578*} &
  \textbf{0.420*} &
   \textbf{0.343*} &
  \textbf{0.234*}&
  0.194* &
  0.132* 
 \\
Ours (N)   & 70.39   & 0.131  & 0.525* & 0.377* & 0.508* & 0.366*& 0.266* & 0.188* & \textbf{0.747*} & \textbf{0.536*}  \\ \hline
\end{tabular}
}
\caption{Evaluation on preference score prediction. Compared with previous works, our predict scores more correctly match the human judgement. We conduct hypothesis test ~\cite{diedenhofen2015cocor}, and * denotes that $p \le 0.01$. }
\vspace*{-1\baselineskip}
\label{tab:main_result}
\end{table*}

\section{Experiments}

\subsection{Compared Methods}
We compare our method with several unreferenced metrics on open story evaluation: Perplexity, Ruber-bert~\cite{ghazarian2019better}, UNION~\cite{guan2020union}, and MANPLTS~\cite{ghazarianplot}.
% To calculate the perplexity of the given story, we use the original GPT2~\cite{radford2019language} (PPL) and the finetuned GPT2 on WP dataset (ft-PPL).
% For Ruber-bert, we finetune the publicly available implementation on WP dataset.
% For UNION and MANPLTS, we use their published pre-trained models.

% We follow ~\citet{guan2021openmeva} to use several unreference metrics as baselines on open story evaluation task.
% Perplexity calculates the probability that a language model can generate the story. We use the original GPT2~\cite{radford2019language} (PPL) and the fine-tuned GPT2 on WP Dataset (ft-PPL) to calculate perplexity of the given story.
% \textbf{RUBER-BERT} is originally proposed in open dialogue generation task and we finetune RUBER-BERT on WritingPrompt Dataset.
% \textbf{UNION} and \textbf{MANPLTS (FT\_LM)} is two neural discriminators that trained from human-written stories and negative samples.

\subsection{Preference Score Evaluation}
\label{sec:overall_evaluation}

\subsubsection{Accuracy and Score Distance}
We evaluate the predicted preference scores obtained by all compared methods on 100k Story Ranking test data. 
Pairwise Ranking Accuracy (Acc) is calculated as the percentage of the story with higher upvotes getting a higher score than the one with lower upvotes.
We also compute the averaged score gap (Dis) between two stories in pairs.
Table~\ref{tab:main_result} (Human (Ranking)) indicates that existing methods on preference-aware story evaluation on human-written stories are close to random selection (i.e., Acc=$0.5$, Dis=$0$).
In contrast, our method can successfully compare two stories and achieve an acceptable score gap between two stories.

% cannot evaluate the stories based on the contents, as their performance are close to random selection (i.e. Acc=$0.5$, Dis=$0$), while our model can successfully compare two stories and achieve an acceptable score gap between two stories. 

\subsubsection{Correlation with Human Judgments}
We calculate the correlation between our predicted preference scores and human judgment for stories. 
We use the correlation metrics Spearman ($\rho$)~\cite{zar1972significance} and Kendall
($\tau$)~\cite{schaeffer1956concerning}, which are known to be beneficial in estimating monotonic associations for not normally distributed and ranked scores.
% \chen{need double check}
We collect and annotate both human-written and machine-generated stories as our test data:

\noindent{\textbf{WP$_{200}$}}.
We collect human judgments for the stories in WP (sampled from test data in Table~\ref{fig:dataset}), where each story is assigned to 8 annotators. 
Annotators are asked to rate each story on a scale of 1 to 5 (from poor to good).
To ensure correctness, we follow~\citet{clarkall} to ask the annotators to compare the stories and write down the reason for clarification. 
We carefully detect the worker behavior and set traps inside the annotation (see Appendix for details).
Finally, we obtain 100 highly-upvoted and 100 lowly-upvoted stories and average the human rates as the target scores in this test data, namely, WP$_{200}$ in the following experiments.
Inside, we witness a higher score for highly-voted stories, proving our hypothesis that upvote counts reflect human preference.

\noindent{\textbf{SCARY$_{200}$}}.
We crawled scary stories from Reddit (r/shortscarystories\footnote{\url{https://www.reddit.com/r/shortscarystories/}}), which are similar to the stories in WP but in a constrained story type. We use the same procedure for WP$_{200}$ to create another human-annotated test dataset, namely SCARY$_{200}$.

\noindent{\textbf{PREF$_{200}$}}.
The same procedure is also used for collecting human annotation for machine-generated stories. 
We select 100 generated stories by LED trained with highly-voted stories in WP and 100 stories by another LED trained with lowly-voted stories. 
We manually ensure that the selected stories do not contain severe coherence issues, and ask the annotators to rate the stories based on whether they enjoy the stories.

\noindent{\textbf{COH$_{200}$}}.
We use the same human collected data in the previous work~\cite{ghazarianplot}\footnote{https://github.com/PlusLabNLP/Plot-guided-Coherence-Evaluation/tree/main/Data/AMT}, which focused on recognizing coherence issues in the machine-generated stories (e.g., repeat plots, conflict logic).

\paragraph{Results.}
Table~\ref{tab:main_result} depicts the correlation between human and automatic evaluation metrics on preference (WP$_{200}$, SCARY$_{200}$ and PREF$_{200}$).
We see that our method outperforms previous methods by a large margin on both human-written and machine-generated stories in terms of human preference. 
% Previous methods are insufficient for evaluating the story on its content, while our model (Ours) outperform these metrics by a large margin on both human-written and machine-generated stories in the evaluation of content.
Not surprisingly, in Table~\ref{tab:main_result} (COH$_{200}$), MANPLTS on story coherence evaluation is against our model, as coherence issue does not frequently happen in our training data (i.e., human-written stories).

We notice that preference-aware judgments (PREF$_{200}$) and coherence-based judgments (COH$_{200}$) are distinct. Metrics that perform well in terms of coherence may perform poorly in terms of preference, and vice versa.
To mitigate the gap between preference and coherence, we train our model using negative stories created by UNION and MANPLTS.
As a result, Ours (N) shows rapidly increasing performance on the evaluation in terms of coherence with a bit of performance drop on the preference-aware evaluation, indicating a potential to take into account both coherence and human preference when evaluating a story.

% \hl{of what? I dont understand the point here}\chen{which part} \hl{from indicating, I dont know what you want to conclude after training model w/ negative samples}\chen{the model w/ negative samples can both considering coherence and content} to evaluate a story in both coherence and content. 

% \begin{table}[t]
% \footnotesize
% \centering
% \resizebox{\linewidth}{!}{
% \begin{tabular}{l|l|l|l|ccccc|cccc}
% \hline
%           &            &            &            & \multicolumn{5}{c|}{Automatic} &  \multicolumn{4}{c}{Human}                           \\
% $o$      & $a$        & $c$        & $N$        & PPL            & BLEU         & ROUGE & METEOR & CIDER & Overall & Rel(story)     & Rel(aspect) & Rel(rating) \\\hline
%           &            & \checkmark &            & 7.31          & 8.45         & 16.63 & 18.81  & 7.39  & 47.61   & \textbf{73.70} & 79.20       & -          \\
% \checkmark &
%   \checkmark &
%   \checkmark &
%   &
%   \textbf{7.06} &
%   \textbf{8.60} &
%   \textbf{16.76} &
%   \textbf{18.88} &
%   \textbf{7.99} &
%   \textbf{49.40} &
%   72.93 &
%   \textbf{82.83} &
%   \textbf{58.33} \\
% \checkmark & \checkmark & \checkmark & \checkmark & 7.95          & 8.36         & 16.69 & 18.44  & 7.07  & 43.45   & 68.64          & 81.84       & 50.49 \\\hline
% \end{tabular}
% }
% \caption{Comment generation evaluation on automatic scores and human evaluation. In human evaluation, the kappa coefficient $\kappa$ for each score are located in 0.4-0.6, indicating a moderate agreement between annotators.}
% \vspace*{-1\baselineskip}
% \label{tab:comment_evaluation}
% \end{table}

\section{Ablation Study}

\subsection{Preference Score Prediction}
In this section, we further test the performance of preference score prediction combined with other components: aspects $\mathbf{a}$, comments $\mathbf{c}$ and negative stories $\mathbf{N}$.
Table~\ref{tab:ablation} summarizes the results by joint training.
When aspects are used, performance decreases in the WP$_{200}$ but increases in the SCARY$_{200}$, and the pattern is reversed when comments are used. 
We also test the model performance trained with the dataset without data augmentation $\triangle$, and we can see that our model trained with augmented data outperforms that with the original data, which shows the significance of data augmentation.

\begin{table}[t]
\centering
% \footnotesize
\resizebox{\linewidth}{!}{
\begin{tabular}{l|l|l|l|cc|cc|cc}
\hline
 &
  &
  &
  &
  \multicolumn{2}{c}{Ranking}&
  \multicolumn{2}{|c}{WP$_{200}$} &
  \multicolumn{2}{|c}{SCARY$_{200}$}  \\
        $p_s$      & $\mathbf{a}$        & $\mathbf{c}$        & $\mathbf{N}$        &    Acc    &    Dis   & $\rho$ & $\tau$ & $\rho$       & $\tau$  \\\hline
\checkmark &            &            &  & 71.10 & 0.212 & 0.557    & 0.401  & 0.533          & 0.389   \\
\checkmark & \checkmark &            &  & 71.99 & 0.221 & 0.525    & 0.378  & \textbf{0.579} & 0.417    \\
\checkmark &            & \checkmark &  & 72.15 & 0.207 & 0.580    & 0.421  & 0.510          & 0.371   \\
\checkmark &
  $\triangle$ &
  $\triangle$ &
  &
  72.95 &
  \textbf{0.229} &
  0.571 &
  0.409 &
  0.564 &
  0.409  \\
\checkmark &
  \checkmark &
  \checkmark &
  &
  \textbf{73.93} &
  0.228 &
  \textbf{0.583} &
  \textbf{0.422} &
  0.578 &
  \textbf{0.420} \\
\checkmark &
  \checkmark &
  \checkmark &
  \checkmark &
  69.02 &
  0.119 &
  0.525 &
  0.377 &
  0.508 &
  0.366 \\\hline
\end{tabular}
}
\caption{Ablation study on preference score prediction. All results are statistical significant $p \textless 0.01$. $\triangle$ means that we use the collected data without augmentation.  More results are listed in supplementary materials.}
% \vspace*{-1\baselineskip}
\label{tab:ablation}
\end{table}

\subsection{Aspect Evaluation}
We evaluate our model for predicting confidence scores and ratings for the aspects.
For confidence scores, we calculate the recall performance on top-k (i.e., k=1,3,5) on the test split of 46K Aspect Rating and Reasoning data to show the percentage of human selected aspects that can be involved within the aspects with top-k confidence.
For ratings, we calculate the correlation between human annotation and our model prediction.
Table~\ref{tab:aspect_evaluation} shows the results compared with joint training other two tasks. Story ranking and reasoning help the model output more correct confidence and ratings.

\begin{table}[t]
\footnotesize
\centering
\resizebox{\linewidth}{!}{
\begin{tabular}{l|l|l|l|ccc|ccc}
\hline
       &        &        &        & \multicolumn{3}{c}{Confidence} & \multicolumn{2}{|c}{Rating}      \\

$p_s$      & $\mathbf{a}$        & $\mathbf{c}$        & $\mathbf{N}$      & R@1      & R@3      & R@5    &  $\rho$             & $\tau$  \\\hline
       & \checkmark &        &        & 16.06    & 46.05    & 73.59    & 0.190       & 0.140   \\
\checkmark & \checkmark &        &        & 17.36    & 51.59    & 76.30    & 0.227*        & 0.168*   \\
\checkmark & \checkmark & \checkmark &  & \textbf{19.94} & \textbf{52.68} & \textbf{79.64} & \textbf{0.248*}  & \textbf{0.185*} \\
\checkmark & \checkmark & \checkmark & \checkmark & 19.88    & 51.44    & 79.20    & 0.216*      & 0.161*  \\\hline
\end{tabular}
}
\caption{Evaluation on aspect confidence and rating. $p_s$, $a$, $c$, $N$ denotes the preference score, aspects, comments and negative samples that are used in training our model respectively. }
\vspace*{-1\baselineskip}
\label{tab:aspect_evaluation}
\end{table}

\subsection{Comment Evaluation}
We evaluate the comment generation with automatic metrics and human evaluation. For automatic scores, we apply Perplexity (PPL), Averaged BLEU1-4 (B), ROUGE (R). 
For human evaluation, we mainly measure the relativeness between comments with the given story \textbf{Rel(s)}, aspect category \textbf{Rel(a)} and rating score (0-1 negative-positive) \textbf{Rel(r)}. We also measure \textbf{Overall (O)} quality by calculating the percentage of the comments that are agreed upon by annotators.
Each comment is assigned to 5 annotators with a binary choice (i.e., related or not related, agree or not agree).
From the result in Table~\ref{tab:comment_evaluation}, 
our generated comments are highly related to the given stories and the aspects. Together with the training on preference score prediction and aspect rating further improve the comment generation performance. The results so far show that the preference score, aspects, and comments all benefit one another, illustrating the significance of incorporating aspects and comments into our task.

\begin{table}[t]
\centering
\resizebox{\linewidth }{!}{
\begin{tabular}{l|l|l|l|ccc|cccc}
\hline
           &            &            &            & \multicolumn{3}{c|}{Automatic} &  \multicolumn{4}{c}{Human}                           \\
$p_s$      & $\mathbf{a}$        & $\mathbf{c}$        & $\mathbf{N}$        & PPL            & B         & R & O & Rel(s)     & Rel(a) & Rel(r) \\\hline
           &            & \checkmark &            & 7.31          & 8.45         & 16.63  & 47.61   & \textbf{73.70} & 79.20       & -          \\
\checkmark &
  \checkmark &
  \checkmark &
   &
  \textbf{7.06} &
  \textbf{8.60} &
  \textbf{16.76}  &
  \textbf{49.40} &
  72.93 &
  \textbf{82.83} &
  \textbf{58.33} \\
\checkmark & \checkmark & \checkmark & \checkmark & 7.95          & 8.36         & 16.69  & 43.45   & 68.64          & 81.84       & 50.49 \\\hline
\end{tabular}
}
\caption{Comment generation evaluation on automatic scores and human evaluation. In human evaluation, the kappa coefficient $\kappa$ for each score are located in 0.4-0.6, indicating a moderate agreement between annotators.}
\vspace*{-1\baselineskip}
\label{tab:comment_evaluation}
\end{table}

\section{Discussion}

\subsection{Pairwise Evaluation with StoryER}
Given a set of prompts, two story generation models can generate stories based on the given prompt.
We have two straightforward ways to compare two models using our proposed preference scores: 1) average the preference scores for stories on each model and compare the mean average scores. 2) perform pairwise comparisons for stories from the same prompt and get the preference percentage.
% As human preference is inevitably influenced by different prompt content, 
We recommend the second method as it strictly follows our pairwise ranking strategy.
% \hl{(give examples on P\&W and w/o)}

\subsection{Domain Transfer in Preference Score}
\label{sec:domain}

To show the generalization of evaluation metrics, we calculate the averaged predicted preference scores for data from different domains (see Table~\ref{tab:other_domain}). 
We compute average scores on 1) lowly-voted (low) and highly-voted stories (high) on both WP$_{200}$ and SCARY$_{200}$, 2) machine-generated stories by LED (LED), and with Plan-and-Write strategy~\cite{yao2019plan} (P\&W) trained separately on the highly-upvoted and lowly-upvoted stories, 3) negative stories generated from previous works~\cite{guan2020union, ghazarianplot}, 4) stories from other datasets: fairy tales (short stories), childbook dataset~\cite{hillgoldilocks} and bookcorpus~\cite{zhu2015aligning}.

As shown in Table~\ref{tab:other_domain}, UNION and MANPLTS consistently produce higher scores for human-written stories (Human and Other blocks) while producing lower scores for machine-generated stories (Machine and N blocks).
While looking into more details, we can see that they cannot successfully distinguish the story quality, e.g., SCARY$_{200}$(low) and SCARY$_{200}$(high) receive identical scores.
These observations strongly indicate that UNION and MANPLTS work well on evaluating coherence but deviate from human preference when evaluating human-written stories.

Our method, on the other hand, is capable of following human preference (Human and Machine block) (also see SCARY$_{200}$(low) and SCARY$_{200}$(high) as an example).
The model trained with highly-voted stories can generate better stories than that trained with lowly-voted stories, and P\&W strategy performs even better as proved in many previous works~\cite{fan2019strategies,tan2021progressive}.
From the results, our model produces higher scores for LED (high) compared with LED (low) and even higher scores for LED P\&W (high), which indicates that our model still follows the human preference on machine-generated stories.
As serious coherence problems do not commonly occur in our training data, our method show failure in recognizing manually created incoherent stories (N block).
However, our model (Ours (N)) works after we incorporate these stories into our training data, leading to a future direction that unifies the coherence-based and preference-aware metrics.
Surprisingly, our model gives relatively low scores when adopting stories from other domains (Other block).
We think this is because the writing style changes the criterion of human preference, which misleads our model to predict a not reasonable score, thus leading us to a big challenge in generalizing preference-aware story evaluation.

\begin{table}[]
\footnotesize
\centering
\resizebox{\linewidth}{!}{
\begin{tabular}{l|l|cc|c|c}
\hline
                                       &                 & \multicolumn{2}{c|}{Coherence}  & Preference  & Hybrid \\
                                       &        Dataset       & UNION & MANPLTS  & Ours  & Ours (N) \\\hline
\multirow{3}{*}{\STAB{\rotatebox[origin=c]{90}{Human}}} & WP$_{200}$(low)     & 0.771 & 0.878 & 0.347 & 0.655                \\
                                         & WP$_{200}$(high)    & 0.837 & 0.948 & 0.692 & 0.884                \\
                                         & SCARY$_{200}$(low)  & 0.833 & 0.825 & 0.355 & 0.625                \\
                                         & SCARY$_{200}$(high) & 0.895 & 0.850 & 0.743 & 0.883        \\\hline
\multirow{3}{*}{\STAB{\rotatebox[origin=c]{90}{Machine}}} & LED (low)       & 0.687 & 0.091 & 0.297 & 0.290                \\
                                         & LED P\&W (low)   & 0.775 & 0.300 & 0.535 & 0.305               \\
                                         & LED (high)       & 0.588 & 0.001 & 0.409 & 0.290               \\
                                         & LED P\&W (high) & 0.760 & 0.393 & 0.573 & 0.308                 \\\hline
\multirow{2}{*}{\STAB{\rotatebox[origin=c]{90}{N}}}         & Negative(UNION)           & 0.360 & 0.003 & 0.244 & 0.019                \\
                                         & Negative(MANPLTS)         & 0.414 & 0.228 & 0.319 & 0.027                \\\hline
\multirow{3}{*}{\STAB{\rotatebox[origin=c]{90}{Other}}}           & fairy tale (short)       & 0.917 & 0.500 & 0.233 & 0.482                \\
                                         & childbook (long)        & 0.886 & 0.915 & 0.318 & 0.476                \\
                                         & bookcorpus (long)       & 0.965 & 0.949 & 0.285 & 0.416              \\\hline 
                                         
\end{tabular}
}
\caption{Our model and existing works on various domains of stories. We report the averaged preference score on stories from four different domains.}
\vspace*{-1\baselineskip}
\label{tab:other_domain}
\end{table}

% \subsection{Limitation}
% Compared with coherence-based story evaluation, preference-aware story evaluation is more sensitive to the domain shift. A detailed analysis is listed in Appendix sec. C. The result indicates that when the story domain, particularly the writing style, varies significantly (e.g., fable and novel), the preference score becomes unreliable. We believe this is because the evaluation criterion for human preference varies according to the story type. This also points to a potential research topic: few-shot learning for preference score prediction on different story types.

\subsection{More Analysis}
Due to the page limit, we put more analysis in the ablation studies. 
In Appendix Sec. D, we witness high correlation scores between preference score and each aspect rating, indicating the effectiveness of all pre-defined aspects in the evaluation.
We also analyze the confidence and rating scores of the horror aspect with the preference score on scary stories in Appendix Sec. E.
The result follows the human intuition that evaluation on scary stories shows a tendency to rely on the horror aspect.

\section{Conclusion} 
In this paper, we investigate a novel task of preference-aware story evaluation, StoryER, which produce a score with explanation through various aspects and comments, bringing gains on both machine-generated and human-written stories evaluation.
To support the task, we present a new dataset consisting of paired ranked stories and more explicit annotation (i.e., rating and reasons) for pre-defined aspects.
Our comprehensive ablation studies and intensive analysis show the effectiveness of using aspect rating and reasoning on preference score prediction.
With the development of story generation, we believe that preference-aware story evaluation will be the mainstream research when machine-generated stories do not suffer from serious coherence problems. Further studies on our dataset can also be conducted to reveal the point that influence the readers to upvote the stories.

\section{Limitations}

Our work (currently) has the following limitations:

\noindent\textbf{(1)} As indicated in Section~\ref{sec:domain}, our proposed metrics are negatively affected by the significant domain shift, since we only take stories from one platform to train our model. Idealistically, a more general model can be trained with all types of stories, but it needs massive annotations on human preference (i.e., upvote counts).

\noindent\textbf{(2)} Since the upvote counts in the original dataset will be influenced by the prompt's topic, typically, fantastic stories get more upvotes than others. Our model is only trained by story pairs within the same topic, thus if a user inputs two unrelated stories, our system will provide unpredictable results. Therefore, we propose using pairwise evaluation with the same given prompt to avoid comparing stories with diverse topics.

\noindent\textbf{(3)} In this work, we propose to implicitly joint training to increase the performance of each task without explicitly addressing the connection of three subtasks. Although we have aspect rating and comment generation, preference score is still the most effective approach to assess the quality of the story. How to use these comments and aspect ratings is a challenge that will be addressed in the future work.

\section{Ethics and Broader Impacts}

We hereby acknowledge that all of the co-authors of this work are aware of the provided \textit{ACM Code of Ethics} and honor the code of conduct.
This work is mainly about propose a novel method in automatic story evaluation.
The followings give the aspects of both our ethical considerations and our potential impacts to the community.

\vspace{.3em}

\paragraph{Dataset.}
We collect the human annotation of the aspect rating and comments via Amazon Mechanical Turk (MTurk) and ensure that all the personal information of the workers involved (e.g., usernames, emails, urls, demographic information, etc.) is discarded in our dataset.
All the stories in our dataset are collected from a public dataset, namely WritingPrompt.
Although we aim at providing a dataset that agreed upon from various people, there might still be unintended biases within the judgements, we make efforts on reducing these biases by collecting diverse comments and replacing the annotators who tends to be racist.

The detailed annotation process (pay per amount of work, guidelines) is included in the appendix and our public website; 
We primarily consider English speaking regions for our annotations as the task requires certain level of English proficiency.

\vspace{.3em}

\paragraph{Techniques.} We benchmark the story evaluation task with conventional metrics and our proposed metric.
As the story evaluation are of our main focus, we do not anticipate production of harmful outputs on our proposed task.

\section{Acknowledgments}

This paper is supported by Institute of AI and Beyond of the University of Tokyo, JSPS KAKENHI Grant Numbers JP19H04166, JP22H00540, 22K17947 and also based on results obtained from a project JPNP20006, commissioned by the New Energy and Industrial Technology Development Organization (NEDO).
For experiments, computational resources of AI Bridging Cloud Infrastructure (ABCI) provided by National Institute of Advanced Industrial Science and Technology (AIST) was used.

\bibliography{anthology,custom}
\bibliographystyle{acl_natbib}

\appendix
\newpage
\section{Website Demo}
\begin{figure*}
    \centering
    \includegraphics[scale=2.5]{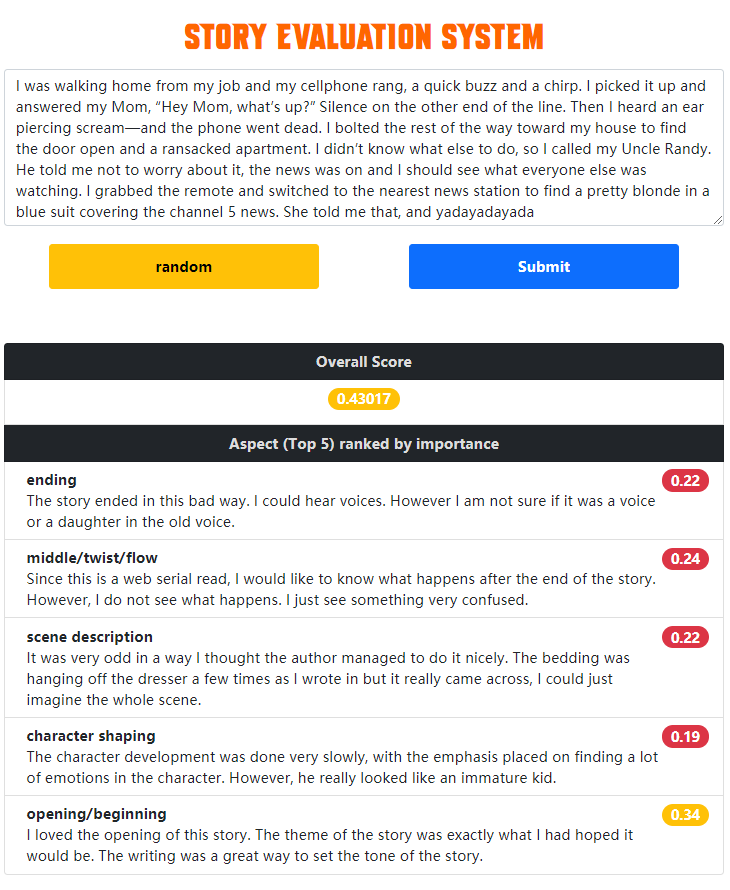}
    \caption{An example on the website.}
    \label{fig:demo}
\end{figure*}

We display the collected data, AMT template and models on our website\footnote{For the review process, dataset and pre-trained model demo are available at anonymous website \url{http://storytelling-lab.com/eval}}.
The users can input their own stories or randomly select one story. The server then runs our model and output a preference score, and comments for each aspect.
Figure~\ref{fig:demo} shows an example.

\section{Code}
We also put our source codes into the supplementary materials. Due to the upload size limitation. We truncate our 100k Story Ranking data into a size of 1000, as well as the 46k Aspect Rating and Reasoning data. Please kindly follow the README to run the experiment.
Our human annotation results can be also found under the folder ``data''.
Additionally, we put some examples for machine-generated stories introduced in our paper.

\section{Correlation Between Story Quality and Aspect Rating}
\begin{figure}
    \centering
    \includegraphics[scale=0.16]{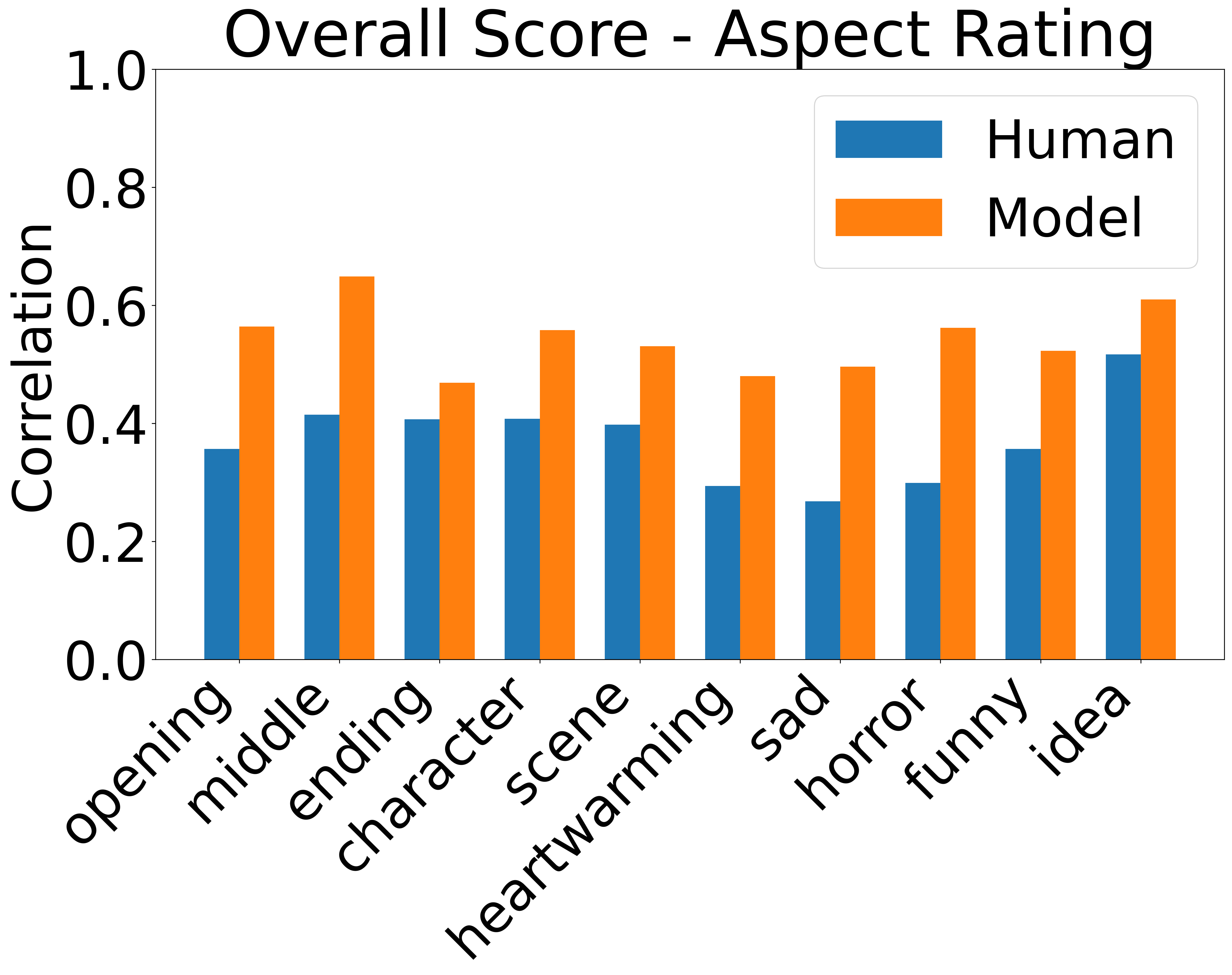}
    \caption{Correlation of upvote number - aspect rating (Human) and the correlation of predicted preference scores and predicted aspect rating (Model (Ours)). The correlation values are all statistical significant. (i.e. $p \le 0.01$)}
    \label{fig:aspect_evaluation}
\end{figure}

We calculate the correlation between human ratings on each aspect with the upvote number, and the predicted aspect rating with the predicted preference score, to figure out the correlation between the aspect rating and the preference score.
The results are listed in Figure~\ref{fig:aspect_evaluation}.
We can see the results from our model greatly match the distribution of the correlation between human aspect rating and human upvote number. None of these shows domination, which proves that all pre-defined aspects affect the final preference score prediction.

\section{Horror/Scary Aspect with SCARY$_{200}$}
To show how aspect ratings and confidence are related to the story, we further analyze their performance on WP$_{200}$ and SCARY$_{200}$. We calculate the recall performance and rating correlation on ``horror/scary'' aspect only to detect how this aspect works in both data.
Table~\ref{tab:horror_cor} depicts that horror aspect can achieve 36\% probability to be the top confident aspect in SCARY$_{200}$, while the number is only 0.5\% in the original WP$_{200}$. On the other hand, the preference score also has a higher correlation with the rating from ``horror/scary'' aspect. 
These results prove that the predicted aspects show high connection to the preference score prediction.

% , due to the page limits, we put the details in the supplementary materials.

\begin{table}[]
\centering
\footnotesize
\resizebox{\linewidth}{!}{
\begin{tabular}{l|ccc|ccc}
\hline
         & \multicolumn{3}{c}{Confidence(Horror)} & \multicolumn{3}{|c}{ Correlation(Horror)} \\
         & R@1         & R@3         & R@5        & $\rho$   & $\tau$ \\\hline
WP$_{200}$       & 0.50        & 11.00       & 13.50      & 0.233         & 0.163    \\
SCARY$_{200}$ & 36.50       & 50.50       & 57.50      & 0.302        & 0.222   \\
\hline
\end{tabular}
}
\caption{Confidence for aspect horror/scary for WP$_{200}$ and SCARY$_{200}$ dataset and the correlation between the preference score and horror/scary aspect ratings in two dataset.}
\vspace*{-1.2\baselineskip}
\label{tab:horror_cor}
\end{table}

\section{Implementation Details}
\subsection{Model for Preference-Aware Story Evaluation}
Our model for story evaluation used pre-trained LED~\cite{beltagylongformer} from  Huggingface\footnote{\url{https://huggingface.co/transformers/model_doc/led.html}}. We finetune the model with 100k Story Ranking data and 46k Aspect Rating and Reasoning data on a machine with 8 NVIDIA A100 GPUs.
During the training, we set the batch size as 16, the margin as 0.3 and run 20k iterations (5 epoch on 100k Story Ranking data) on training (10 hours).
In each iteration, we adopt two pairs: one from 100k Story Ranking data and the other from 46k Aspect Rating and Reasoning data, to our model.
We take AdamW optimizer~\cite{loshchilov2018decoupled} with an initial learning rate of 4e-6, warming up in the first epoch and decreasing by a linear schedule. The reported results are averaged by the best results from three models with the same structure but initialized with three different seeds. 
For hyper-parameter search, we search margin $m$ from 0.2 to 1.0 with the step of 0.1, learning rate from 4e-4, 4e-5, 4e-6 and 4e-7, and record the best hyper-parameters.

\subsection{Model for Aspect Category Classification}
\label{sec:aspect_classification}
Our model for aspect category classification is based on RoBERTa large~\cite{liu2019roberta}. Same as the model we used in preference score prediction, we apply a linear projection on the feature of [CLS], the first token of the input comments. We then train the model with cross-entropy loss for 20 epochs with the learning rate 4e-5.

\subsection{Model for Comment Sentiment Analysis}
\label{sec:aspect_sentiment}
Our model for comment sentiment analysis uses the same model structure for aspect category classification. We also use the same epochs number and learning rate during the training. The only difference is that the targets in training are the sentiment rate with a scale of 1-5 (from definitely negative to definitely positive)

\section{More Results}

\subsection{Ranking vs Discrimination}
\label{sec:rank}
% \hl{need revision}
Given two types of stories, highly and lowly upvoted, a straightforward method to build the model is through discrimination; use 0 and 1 as target with cross-entropy loss.
We compare the results by using ranking and discrimination. The result is shown Table~\ref{tab:rankvsdis}. 
From the result, we see that ranking strategy achieves better scores than discrimination and that with label smoothing. We believe it is because when we conduct ranking, we only enlarge the preference scores between stories written from the same prompt. The encoder can learn better how human preference works by comparing stories with the same topic. 
On the other hand, the ranking loss is more flexible compared with binary classification, which can be easily extended to rank more than two types of stories as shown in Equation~\ref{equ:neg}.

\begin{table}[tb]
\footnotesize
\centering
\resizebox{\linewidth}{!}{
\begin{tabular}{l|c|cc|cc}
\hline
               & \multicolumn{1}{c|}{Ranking} & \multicolumn{2}{c|}{WP$_{200}$} & \multicolumn{2}{c}{SCARY$_{200}$}                              \\
               & Acc                  & $\rho$         & $\tau$        & $\rho$ & $\tau$\\ \hline
CE  & 72.82              & 0.539          & 0.390         & 0.538  & 0.412   \\
CE(smooth) & 71.38            & 0.550          & 0.394         & 0.561  & 0.414   \\
Ranking &
  \textbf{73.93} &
  \textbf{0.583} &
  \textbf{0.422} &
  \textbf{0.578} &
  \textbf{0.420} \\\hline
\end{tabular}
}
\caption{Comparison of discrimination and ranking.}
\label{tab:rankvsdis}
\vspace*{-1\baselineskip}
\end{table}

\subsection{PPL in automatic story evaluation}
For an interesting finding, Perplexity (PPL) shows positively correlated to the score of WP and more highly correlated to the score of SCARY$_{200}$, while showing substantially negatively correlated to the score of coherence on machine-generated stories, which reveals a potential for story evaluation using pre-trained language models.

\subsection{Results of Comment Evaluation}
Due to the page limitation, we put the results of comment evaluation with more metrics in Table~\ref{tab:comment_evaluation}. We see that our model achieves higher performance on most of the metrics.

\begin{table*}[t]
\footnotesize
\centering
\resizebox{\linewidth}{!}{
\begin{tabular}{l|l|l|l|ccccc|cccc}
\hline
          &            &            &            & \multicolumn{5}{c|}{Automatic} &  \multicolumn{4}{c}{Human}                           \\
$p_s$      & $a$        & $c$        & $N$        & PPL            & BLEU         & ROUGE & METEOR & CIDER & Overall & Rel(story)     & Rel(aspect) & Rel(rating) \\\hline
          &            & \checkmark &            & 7.31          & 8.45         & 16.63 & 18.81  & 7.39  & 47.61   & \textbf{73.70} & 79.20       & -          \\
\checkmark &
  \checkmark &
  \checkmark &
  &
  \textbf{7.06} &
  \textbf{8.60} &
  \textbf{16.76} &
  \textbf{18.88} &
  \textbf{7.99} &
  \textbf{49.40} &
  72.93 &
  \textbf{82.83} &
  \textbf{58.33} \\
\checkmark & \checkmark & \checkmark & \checkmark & 7.95          & 8.36         & 16.69 & 18.44  & 7.07  & 43.45   & 68.64          & 81.84       & 50.49 \\\hline
\end{tabular}
}
\caption{Comment generation evaluation on automatic scores and human evaluation. In human evaluation, the kappa coefficient $\kappa$ for each score are located in 0.4-0.6, indicating a moderate agreement between annotators.}
\vspace*{-1\baselineskip}
\label{tab:comment_evaluation}
\end{table*}

\subsection{Results of Aspect Category Classification}
\begin{figure}[h!]
    \centering
    \includegraphics[scale=0.5]{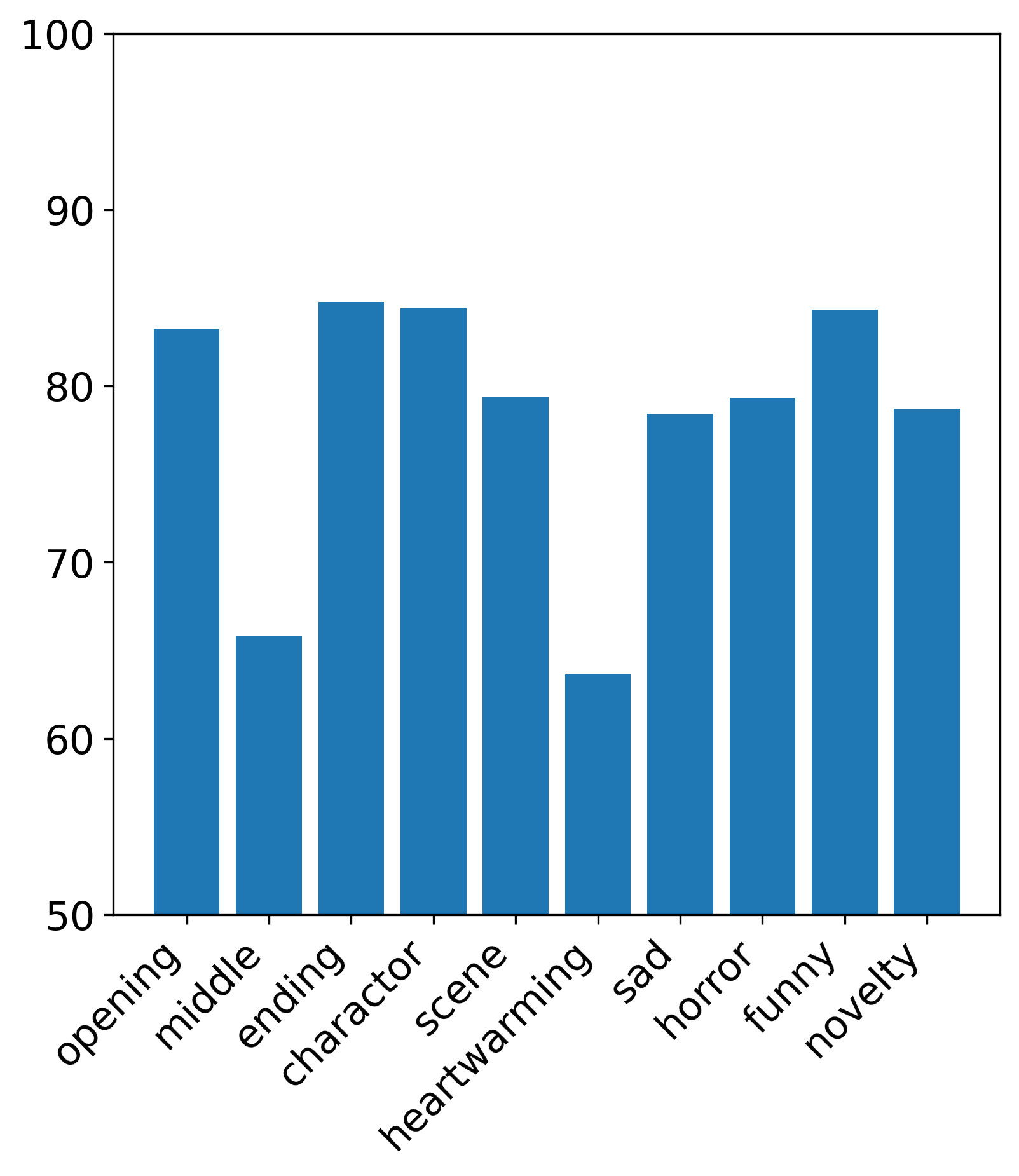}
    \caption{Comment classification results.}
    \label{fig:aspect_class}
\end{figure}

We use aspect category classification model, introduced in Sec.~\ref{sec:aspect_classification}, for filtering out noisy comments.
Figure~\ref{fig:aspect_class} shows the classification results. Except for ``ending'' and ``heartwarming'', all aspect classes can achieve an average of around 80\% accuracy, showing high performance on classification. 
We filter out the comments, with no aspect category score exceeding 0.9 after softmax function.

\subsection{Results of Comment Sentiment Analysis}
Comment Sentiment Analysis model, introduced in Sec.~\ref{sec:aspect_sentiment}, is used to rate comments by their sentiments.
Table~\ref{tab:comment_senti} shows the results. Our output is the rates from 1 to 5. In the evaluation, we simply group 1 and 2 as the negative, 3 as the neutral, 4 and 5 as the positive. 
The results show that our sentiment analysis model can correctly predict the sentiment, especially on positive and negative.

\begin{table}[h!]
\centering
\resizebox{\linewidth}{!}{
\begin{tabular}{l|llll}
\hline
    & positive & neutral & negative & average \\\hline
Acc & 89.70\%   & 50.93\%  & 85.20\%   & 83.03\% \\\hline
\end{tabular}
}
\caption{Comment sentiment analysis results}
\label{tab:comment_senti}
\end{table}

\subsection{Comment Data Augmentation}
We collect over 150k uncategorized comments from metadata in WP. We use the aspect category classification model and filter out the irrelevent comments. However, we found bias inside the comments. For example, we get almost 9000 comments about ``ending'', while only 1200 for ``sad''.
To mitigate the bias that would be inducted into our story evaluation model, we sample about 2000 comments for each aspect, and use all comments for the aspect which contains less than 2000 comments. The final data statistics of comments can be referred to our website.

\section{Human Annotation}
\subsection{Human Annotation on Test Data}
For evaluation, we collect human judgments through AMT for 200 highly-upvoted stories and 200 lowly-upvoted stories from WP (sampled from test data in 100k Story Ranking data), where each story is assigned to 8 annotators. 
Annotators are asked to rate each story on a scale of 1 to 5 (from poor to good).
Following~\citet{clarkall}, we asked the annotators to compare the stories before rating and write down a very brief reason for clarification. 
To further ensure the correctness of the annotation, we calculate the statistics of the annotator behavior (i.e., working time per hit) and set traps in the batch (i.e., insert extremely poor story, duplicate stories for one annotator to test their consistency).
The submissions from annotators with poor quality are all rejected and then recollected from new annotators. 
Finally, we exclusively keep the 100 highly-upvoted and 100 lowly-upvoted stories with the lowest variance from 8 annotators and average the human rates as the target scores in this test data, namely, WP$_{200}$ in the following experiments. Annotators get \$0.2 as the reward for each submission.
Besides, we crawled scary stories from Reddit (r/shortscarystories~\footnote{\url{https://www.reddit.com/r/shortscarystories/}}), which have a similar writing style to the stories in WP but in a constrained story type. We repeat the procedure for WP$_{200}$ and create another human-annotated test data, namely SCARY$_{200}$. 
The same procedure is also used for collecting human annotation on machine-generated stories. We generate 200 stories using LED trained with highly-voted stories and another 200 stories using LED trained with lowly-voted stories for annotation. We ask the annotators to rate the stories based on human preference and also ask them to distinguish whether the given stories are human-written or machine-generated. We exclusively keep the stories that deceive the annotators, as these stories do not contain serious coherence problems.

\subsection{Data Collection}
In this paper, we mainly collect data for two different uses.
% \noindent \textbf{StoR3} is a novel dataset we collected for preference-aware story evaluation.
\noindent Annotators get \$1 as the reward for each submission. The total data collection takes 2 months. To assess the quality of each annotator, we randomly sample the submissions from each annotator every two days, bonus the one with good quality and warn the annotators who give nonsense comments.

\subsection{Human Annotation Inner-Agreement}
As we assign one story for more than one annotator, we calculate the inner-agreement from different annotators on aspect selection. 
As a result, 65.80\% aspects are selected by more than one annotator, and the correlation coefficient of multi-annotation on aspect ratings are 0.913 and 0.811, corresponding to the Spearman~\cite{zar1972significance} and Kendall~\cite{schaeffer1956concerning} respectively.

\section{Aspect Category Name Definition}
As no standard criterion exists for story evaluation, we collect some well-used aspects that used in the Internet. We mainly refer to the websites~\footnote{\url{https://en.wikipedia.org/wiki/List_of_writing_genres}}~\footnote{\url{https://www.writerswrite.co.za/the-complete-guide-to-evaluating-your-short-story/}}~\footnote{\url{https://www.oprahdaily.com/entertainment/books/a29576863/types-of-book-genres/}}.

\end{document}